%% file: main.tex
\documentclass{article}

\PassOptionsToPackage{numbers, compress}{natbib}
\PassOptionsToPackage{table, dvipsnames}{xcolor}

\usepackage[main, final]{neurips_2026}

\usepackage[utf8]{inputenc}
\usepackage[T1]{fontenc}
\usepackage{hyperref}
\usepackage{url}
\usepackage{booktabs}
\usepackage{amsfonts}
\usepackage{nicefrac}
\usepackage{microtype}
\usepackage{xcolor}

\input{preamble}

\title{ChanSFormer: A Channel Agnostic Vision Transformer for Multi-Channel Cell Painting Images}

\author{%
  Jingwei Zhang\\
  Stony Brook University \\ 
  \texttt{jingwezhang@cs.stonybrook.edu} \\
  \And
  Srinivasan Sivanandan \\ 
  Insitro \\
  \texttt{srinivasan@insitro.com}
  \And
  Dimitris Samaras \\ 
  Stony Brook University \\
  \texttt{samaras@cs.stonybrook.edu}
}

\begin{document}

\maketitle

\input{0_abstract}
\input{1_intro}
\input{2_related}
\input{3_method}
\input{4_experiments}
\input{5_conclusion}

\begin{ack}
This work builds upon a preliminary version presented at the ImageOmics workshop at NeurIPS 2025, which was conducted while Jingwei Zhang was interning at Insitro. The subsequent extensions and analyses at Stony Brook University were supported by USA NSF grants IIS-2123920 \CM{[D.S.]}, IIS-2212046 \CM{[D.S.]}.

Disclosure of Funding and Competing Interests: Srinivasan Sivanandan is employed by Insitro and may hold equity in the company. Insitro has filed a patent application related to the preliminary work with Jingwei Zhang and Srinivasan Sivanandan listed as inventors. The remaining author declares no competing financial interests.

\end{ack}

{
\small
\bibliographystyle{ieeetr}
\bibliography{bib}
}

\input{6_supp}

\end{document}

%% file: preamble.tex
\usepackage{multirow}
\usepackage{bm}
\usepackage{placeins}
\usepackage{amsmath}
\usepackage{graphicx}
\usepackage{subcaption}
\usepackage{enumitem}

\newcommand{\CM}[1]{#1}
\newcommand{\CMcolor}{}
\newcommand{\CRdag}{\rlap{\CM{$^\dagger$}}}

%% file: 0_abstract.tex
\begin{abstract}
High-content multi-channel imaging, from Cell Painting assays to remote sensing, is central to modern scientific pipelines. Because these channels are highly heterogeneous and experimental configurations frequently evolve, a practical vision backbone must be fundamentally channel-agnostic. Current channel-adaptive vision transformers rely on global self-attention with learnable channel embeddings, which can be computed only for previously known channels. Attempts to overcome this limitation of new channels either discard the embeddings, losing channel identity, or process each channel independently, thereby eliminating critical cross-channel interactions. 
To address this, we introduce \textbf{ChanSFormer}, an efficient Vision Transformer that replaces rigid embeddings with disentangled spatial-channel attention and individual channel CLS tokens. This architecture natively preserves both distinct channel identities and cross-channel information flow. Furthermore, the individual channel class token design unlocks more informative feature representations for self-supervised learning and enables representation-based channel sampling while remaining channel-agnostic. 
We evaluate ChanSFormer on two biology multi-channel datasets, CHAMMI, JUMP-CP and a satellite dataset, So2Sat. Experimental results show that ChanSFormer outperforms state-of-the-art methods by \CM{3.93\%--4.27\% on JUMP-CP} in classification accuracy. It also demonstrates exceptional cross-dataset transferability under a self-supervised setting, outperforming previous methods by up to 2.72\% in accuracy in the same classification tasks. Furthermore, the disentangled attention reduces the quadratic complexity of channel $\times$ spatial sequence length of the global attention baseline, thus improving throughput by 55\%--313\%. \CM{Code is available at \url{https://github.com/cvlab-stonybrook/ChanSFormer}.}
\end{abstract}

%% file: 1_intro.tex
\section{Introduction}
\label{sec:intro}
High-content imaging using the Cell Painting assay is a cornerstone of modern drug discovery~\cite{von2021democratising,vincent2022phenotypic,boutros2015microscopy}. This technique generates images with multiple channels, typically five or more, where each channel uses a specific fluorescent dye to reveal a distinct cellular component such as the nucleus, mitochondria, or cytoskeleton. Because every channel carries a different layer of biological information, the resulting data is highly heterogeneous across channels~\cite{bray2016cell}. Compounding this, research pipelines often evolve through experiments that use varying numbers of channels or introduce novel fluorescent markers. A practical vision backbone for this setting must therefore be flexible and channel-agnostic, i.e., invariant to the number and ordering of input channels.

Vision Transformers (ViTs)~\cite{dosovitskiy2020image_vit} help analyze these complex images, but they still have their limitations. Given the heterogeneous nature of the channels, standard ViTs on cell images tokenize each channel independently~\cite{bao2023channel_cvit,bourriez2024chada_chadavit,pham2024enhancing_dichavit,kraus2024masked_camae} and then employ a single, global self-attention mechanism that processes the concatenated tokens from all channels. To distinguish between tokens originating from different channels within this shared computational space, a learnable channel embedding is added to each token~\cite{bao2023channel_cvit,pham2024enhancing_dichavit}. Although this solves the channel-identity problem, it introduces a critical limitation: the model becomes channel-specific, hard-coded to the initial channel configuration and unable to adapt to experimental setups with potential new channels. 

Some methods \cite{kraus2024masked_camae,de2025scaling} try to address this channel-specificity by modifying the transformer architecture. However, they either drop the channel embedding \cite{kraus2024masked_camae}, so that global self-attention can no longer distinguish tokens from different channels, or embed each channel independently \cite{de2025scaling}, so that the backbone never models cross-channel interactions. 

Another limitation of the global-attention template is computational. Concatenating tokens from all channels dramatically increases the input sequence length: a $224 \times 224$ image with 8 channels yields 8 times more tokens than a natural RGB image at the same resolution. This exacerbates the quadratic complexity of self-attention and slows both training and inference~\cite{bao2023channel_cvit}. 

To address these limitations, we introduce \textbf{ChanSFormer}, a Vision Transformer designed from the outset to be fundamentally channel-agnostic. Our main contributions are:
\begin{itemize}[leftmargin=*]
    \item \textbf{Disentangled channel-spatial attention.} We replace monolithic global attention with separate channel attention and spatial attention. Channel attention natively preserves cross-channel interactions without requiring hard-coded channel embeddings. 
    \item \textbf{Individual channel CLS tokens.} We introduce individual channel CLS tokens that dynamically carry distinct channel information by copying a single learnable CLS token to each channel before the first layer and then allowing the tokens to evolve in later layers through the disentangled attention. This individual token design also unlocks a more informative representation for self-supervised learning.
    \item \textbf{Higher throughput.} The disentangled attention reduces the quadratic complexity of channel $\times$ spatial sequence length in the global attention to the quadratic complexity of channel length plus the quadratic complexity of spatial length for faster training and inference.
\end{itemize}

Extensive experiments in both supervised and self-supervised (DINOv2 \cite{oquab2023dinov2}) regimes validate our approach on two biology multi-channel datasets, CHAMMI, JUMP-CP and a satellite dataset, So2Sat. ChanSFormer achieves comparable or superior accuracy compared to the state-of-the-art (SOTA) methods~\cite{pham2024enhancing_dichavit,bao2023channel_cvit} in the supervised classification, outperforming them by \CM{3.93\%--4.27\% on JUMP-CP} in classification accuracy. Moreover, it demonstrates exceptional cross-dataset transferability under a self-supervised protocol, outperforming SOTA methods by up to 2.72\% in accuracy in the same classification tasks. Finally, the disentangled attention accelerates throughput by {55\%--313\%} compared to the global-attention baseline.

%% file: 2_related.tex
\section{Related work}
\label{sec:related}

\paragraph{Channel-adaptive models for multi-channel imaging.}
Multi-channel microscopy and remote-sensing data have channels that differ in physical meaning across experiments, so a model that commits to a fixed channel ordering at input time is of limited use. The CHAMMI benchmark~\cite{chen2023chammi_depthvit} framed this problem for cell imaging and introduced ConvNeXt or ViT backed channel-adaptive baselines, DepthwiseViT, TemplateViT~\cite{plummerNPAS2022,savarese2018learning}, and HyperViT~\cite{haHypernetworks2016}, that each replace only the first layer to accept variable channel counts while sharing the rest of the backbone. Vision Transformer approaches keep the backbone unified throughout the depth: ChannelViT~\cite{bao2023channel_cvit} tokenizes each channel with a shared linear projection, adds a learnable \emph{channel embedding} to every token, and runs global self-attention over the concatenated sequence. It also introduces Hierarchical Channel Sampling (HCS) that samples channels uniformly at random during training, to improve its channel robustness. ChAda-ViT~\cite{bourriez2024chada_chadavit} follows the same global-attention template while padding inputs to a fixed channel count; DiChaViT~\cite{pham2024enhancing_dichavit} extends this design with a diverse channel sampling strategy and auxiliary losses that push the learned channel embeddings apart. These ViT-based methods share a single architectural choice: a global self-attention that relies on the channel embedding to distinguish tokens from different channels. That choice ties the trained weights to a specific channel configuration.

\paragraph{Self-supervised pretraining for multichannel imaging.}
Self-supervised pretraining for natural images has converged on joint-embedding frameworks such as DINO~\cite{caron2021emerging}, iBOT~\cite{zhou2021ibot}, and DINOv2~\cite{oquab2023dinov2}, and on masked-image modelling as in MAE~\cite{he2022masked}. Extending these frameworks to multichannel imaging has followed several directions. Cell-DINO~\cite{moutakanni2025cell} applies DINOv2 to cellular microscopy with channel-aware augmentations but treats the channel stack as fixed. ChA-MAEViT~\cite{phamcha} pairs masked autoencoding with a channel-aware multi-channel ViT, targeting cross-channel learning through the MAE objective. CA-MAE~\cite{kraus2024masked_camae} also uses masked autoencoding but removes the channel embedding, giving an architecturally flexible ViT that accepts any channel subset at inference. BoC~\cite{de2025scaling} pretrains a microscopy foundation model at scale with DINOv2 by running the backbone on each channel independently. IC-ViT~\cite{lian2025isolated} takes a two-stage route that uses masked-image pretraining on single channels followed by multi-channel finetuning, so cross-channel interactions are learned only at finetuning time.

To achieve channel agnosticism, prior methods adopt two opposing extremes. CA-MAE~\cite{kraus2024masked_camae} removes channel embeddings while retaining global attention, severely compromising the model's ability to distinguish distinct channels. Conversely, BoC~\cite{de2025scaling} processes each channel entirely independently, preserving channel identity but eliminating critical cross-channel information flow during feature extraction. ChanSFormer bridges this gap: by employing decomposed channel-spatial attention alongside individual channel CLS tokens, our architecture natively preserves both continuous cross-channel interactions and distinct channel identities without relying on fixed channel embeddings. \CM{Our decomposition is related to factorized attention in video~\cite{bertasius2021space_timesformer} and multi-view~\cite{miao2025multiview} models, which encode each frame or view with shared transformer blocks and interleave attention across frames or views. Unlike frames, which are ordered in time, or views, which are related by camera geometry, microscopy channels are pixel-aligned but unordered; ChanSFormer therefore encodes no channel ordering, making the encoder permutation-equivariant along the channel axis (Appendix~\ref{sec:supp:factorized}).}

%% file: 3_method.tex
\section{Method}
\label{sec:method}

We present {ChanSFormer}, a Vision Transformer \cite{dosovitskiy2020image_vit} designed to be channel-agnostic for multichannel imagery while remaining efficient and accurate. The model operates on an input $X \in \mathbb{R}^{C \times H \times W}$, where $H$ and $W$ are the spatial dimensions and $C$ is the number of channels.

\subsection{Overall architecture}
\label{sec:method:arch}

\begin{figure}[t]
  \centering
  \includegraphics[width=1.0\linewidth]{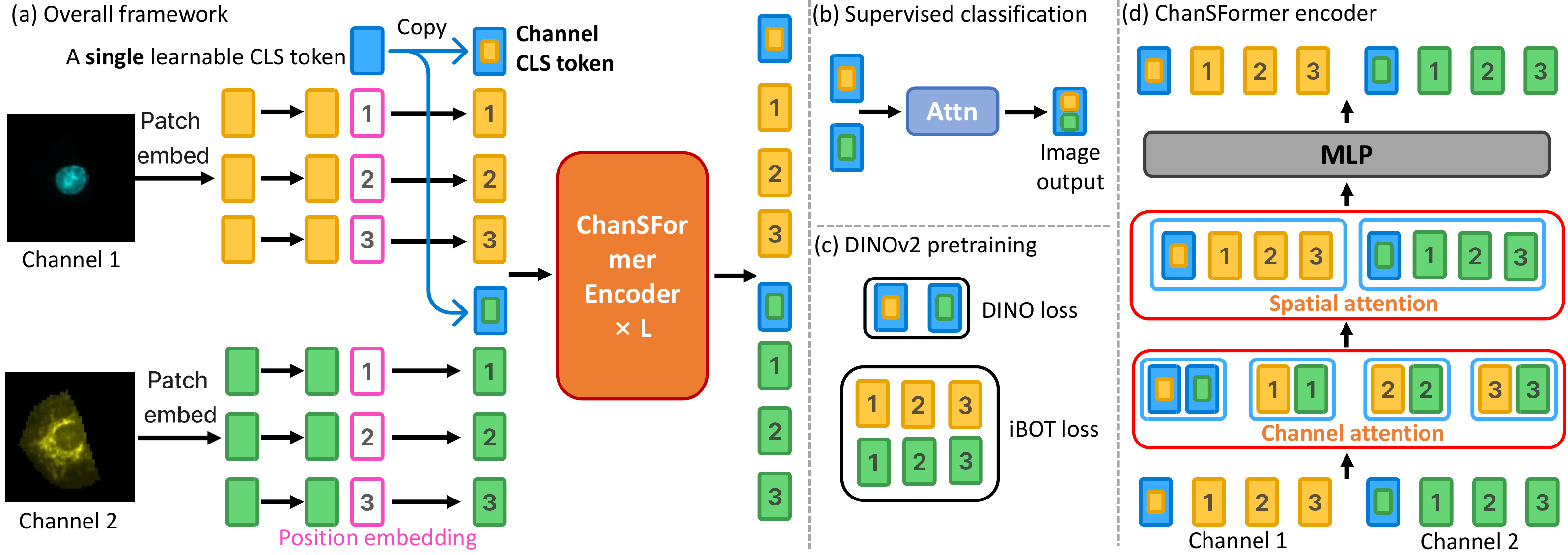}
  \caption{\textbf{(a) Overall framework of ChanSFormer}. Each channel of an input image is independently projected by a shared patch embedding; a shared spatial positional embedding is added, and individual channel class (CLS) tokens, copied from a single learnable CLS token, are concatenated. The resulting tokens pass through L ChanSFormer encoders. \textbf{(b) Supervised classification}. The channel CLS tokens are aggregated via attention into an image-level representation. \textbf{(c) Self-supervised DINOv2 pretraining}. The channel CLS tokens are treated as distinct samples to generate the DINO loss \cite{oquab2023dinov2}, while the remaining patch tokens generate the iBOT \cite{zhou2021ibot} loss. \textbf{(d) Architecture of the ChanSFormer encoder}. A single global attention block is replaced by channel attention (across channels, per spatial position) followed by spatial attention (across spatial positions, per channel).
  }
  \label{fig:framework}
\end{figure}

Figure~\ref{fig:framework} (a) gives an overview, each of the $C$ input channels is independently processed by a shared patch-embedding layer $\Phi : \mathbb{R}^{P \times P} \to \mathbb{R}^{D}$ with patch size $P$ and embedding dimension $D$, producing $N = (H/P)(W/P)$ spatial tokens per channel. A spatial positional embedding in $\mathbb{R}^{N \times D}$ is added, which is \emph{shared} across channels. Individual channel class (CLS) tokens (Section~\ref{sec:method:cls}) are prepended to each channel's tokens, producing the encoder input $z^{(0)} \in \mathbb{R}^{C \times (N+1) \times D}$.

This encoder input $z^{(0)}$ is then fed into a stack of $L$ ChanSFormer encoders, each of which decomposes the global self-attention of a standard ViT encoder~\cite{dosovitskiy2020image_vit} into a channel and a spatial attention (Section~\ref{sec:method:attn}). The output of the stack is $z^{(L)} \in \mathbb{R}^{C \times (N+1) \times D}$; we extract its $C$ channel CLS tokens $\{\mathrm{CLS}_c^{(L)}\}_{c=1}^{C}$ as the per-channel representation. A dedicated channel sampling mechanism (Section~\ref{sec:method:sampling}) varies the number and identity of input channels during supervised training.

Two output heads consume $\{\mathrm{CLS}_c^{(L)}\}$ depending on the training objective.

\textbf{Supervised aggregation.}
As shown in Figure \ref{fig:framework}(b), for supervised classification tasks, we aggregate the channel CLS tokens into a single image token $z_{\text{img}}$ via a query-based attention pooling:
\begin{equation}
z_{\text{img}} = \mathrm{softmax}\!\left( {q\, K^{\text{pool}}\bigl(\mathrm{CLS}^{(L)}\bigr)^{\top}}/{\sqrt{D}} \right) V^{\text{pool}}\bigl(\mathrm{CLS}^{(L)}\bigr),
\label{eq:aggregator}
\end{equation}
where $q \in \mathbb{R}^{D}$ is a single learnable query, $K^{\text{pool}}$ and $V^{\text{pool}}$ are learnable linear projections from $\mathbb{R}^{D}$ to $\mathbb{R}^{D}$, and $\mathrm{CLS}^{(L)} \in \mathbb{R}^{C \times D}$ stacks the channel CLS tokens produced by the last encoder. A single query keeps the pool cost linear in $C$.

\textbf{Self-supervised adaptation.}
As shown in Figure \ref{fig:framework}(c), for DINOv2~\cite{oquab2023dinov2} self supervised pretraining, we do not use an aggregator but feed all the channel CLS tokens directly to the student and teacher heads, treating $\{\mathrm{CLS}_c^{(L)}\}_{c=1}^{C}$ as $C$ independent views of the input. The remaining patch tokens are used to generate the iBOT~\cite{zhou2021ibot} loss. This setup is closely related to BoC~\cite{de2025scaling}, where each channel is independently embedded, except that our ChanSFormer encoder still allows cross-channel information exchange inside the backbone. Using $C$ channel CLS tokens rather than a single aggregated token also yields more supervision per image for the self-distillation objective, increasing the effective batch size and the diversity within the batch. 

\subsection{Individual channel class tokens}
\label{sec:method:cls}

The first important design of ChanSFormer is the individual channel class token.
In standard ViT, a class token is used to aggregate information across spatial positions. For multi-channel images, ideally each channel should have its own learnable class token.
However, to keep the model channel-agnostic, we instantiate a single learnable CLS token $\mathrm{CLS} \in \mathbb{R}^{D}$ and \emph{copy} it into $C$ channels at the input, so that $\mathrm{CLS}_c^{(0)} = \mathrm{CLS}$ for all $c$.
The tokens are initially identical, but because each $\mathrm{CLS}_c$ interacts only with its own channel's spatial tokens during spatial attention (Section~\ref{sec:method:attn}), they evolve through the stack and come to represent features of one channel each.

This departs from the convention in TimeSformer~\cite{bertasius2021space_timesformer}, which pools or averages a single global CLS across time frames. Averaging is appropriate when slices carry highly correlated information (e.g., frames of a video or RGB channels) but is lossy when each slice is semantically distinct, as in multichannel fluorescence microscopy where different channels image different cellular components. Keeping one CLS per channel preserves the channel-specific information for the downstream heads described in Section~\ref{sec:method:arch}.

Apart from its primary role as a channel-agnostic aggregator, the channel CLS token has two other important uses. 
First, in self-supervised learning, as mentioned in Section~\ref{sec:method:arch}, it serves as an independent image view for each channel, effectively increasing the batch size and within-batch diversity.
Second, in supervised learning, it allows us to sample channels according to the relationships among these channel CLS tokens, which we detail in Section~\ref{sec:method:sampling} to drive channel sampling. This sampling role is analogous to the learnable channel embedding used in channel-specific ViTs~\cite{pham2024enhancing_dichavit}.

\subsection{Disentangled channel-spatial attention}
\label{sec:method:attn}

As shown in Figure \ref{fig:framework} (d), each ChanSFormer encoder $\ell$ replaces the multi-head self-attention block in a standard ViT~\cite{dosovitskiy2020image_vit} with two attentions applied sequentially: first a \emph{channel attention}, then a \emph{spatial attention}. The rest of the block (MLP, layer normalization, residual structure) remain unchanged. Let $z^{(\ell-1)} \in \mathbb{R}^{C \times (N+1) \times D}$ be the input.

\textbf{Channel attention.}
At each spatial position $s \in \{1, \ldots, N+1\}$, we attend across the $C$ channel tokens. Let $z^{(\ell-1)}_s \in \mathbb{R}^{C \times D}$ denote the tokens at position $s$, the channel-attention is
\begin{equation}
z'_s = z^{(\ell-1)}_s + \mathrm{CAttn}\!\left(\mathcal{LN}\bigl(z^{(\ell-1)}_s\bigr)\right),
\qquad
\mathrm{CAttn}(x) = \mathrm{softmax}\!\left(\frac{Q'(x)\, K'(x)^{\top}}{\sqrt{D}}\right) V'(x),
\label{eq:channel_attn}
\end{equation}
with $Q', K', V'$ learnable linear projections and $\mathcal{LN}$ denoting layer normalization. Information crosses the channel axis but not the spatial axis at this step.

\textbf{Spatial attention.}
Then, for each channel $c \in \{1, \ldots, C\}$, we attend across the $(N+1)$ spatial tokens of that channel. Let $z'_c \in \mathbb{R}^{(N+1) \times D}$ denote the tokens for channel $c$, the spatial-attention is 
\begin{equation}
z''_c = z'_c + \mathrm{SAttn}\!\left(\mathcal{LN}\bigl(z'_c\bigr)\right),
\qquad
\mathrm{SAttn}(x) = \mathrm{softmax}\!\left(\frac{Q(x)\, K(x)^{\top}}{\sqrt{D}}\right) V(x),
\label{eq:spatial_attn}
\end{equation}
with $Q, K, V$ learnable linear projections from $\mathbb{R}^{D}$ to $\mathbb{R}^{D}$ shared across all channels. Compared with a global self-attention over $CN$ tokens, one ChanSFormer encoder costs $O(CN^2 + NC^2)$ instead of $O((CN)^2)$. The saving is pronounced when $N$ is large (small patch size) or $C$ is moderate, the regime typical of multichannel microscopy.

\textbf{Hardware-aware channel attention optimization.}
Channel attention operates in a regime that breaks the assumptions of standard attention implementations. Modern attention algorithms, including Memory Efficient Attention~\cite{xFormers2022}, assume a high \textit{hardware occupancy} originating from long input sequence length, such as $N$ in spatial attention or $CN$ in global attention. However, channel attention operates over a sequence length of $C$ (typically $< 16$), which is too short to saturate NVIDIA TensorCores for matrix multiplication. Meanwhile, existing algorithms require the sequence length dimension at a fixed location. Thus, channel attention must conduct excessive tensor permutations to meet this requirement. As a result, when applied trivially to channel attention, existing algorithms will suffer from degraded throughput due to \textit{\textbf{low hardware occupancy}} and \textit{\textbf{excessive memory operations}}. 

To resolve these problems, we fuse the entire channel attention computation into a single Triton kernel and batch multiple short instances together, so that the memory overheads are amortized across hundreds of attention operations rather than paid per instance. Furthermore, our attention kernel adapts its internal computation to the channel size $C$. At small sizes that are unable to saturate TensorCores, we fall back to CUDA cores for a higher throughput (see Table~\ref{tab:speed_kernel}); at larger sizes, we engage the TensorCores for matrix multiplication, as in regular attention algorithms. These novel design choices place our kernel in a complementary regime to FlashAttention~\cite{dao2022flashattention,dao2023flashattention}: where FlashAttention optimizes a moderate number of long sequences, ours optimizes an excessive number of short sequences.

\subsection{Representation based channel sampling}
\label{sec:method:sampling}

Randomly sampling the input channel set during training is an effective regularizer for channel-adaptive ViTs~\cite{bao2023channel_cvit, pham2024enhancing_dichavit}, improving models' robustness. DiChaViT~\cite{pham2024enhancing_dichavit} demonstrated that guiding this sampling using the pairwise distances among learned channel embeddings outperforms uniform random sampling (HCS \cite{bao2023channel_cvit}). However, this strategy tightly couples the sampler to the channel embedding, the very component we removed to make ChanSFormer channel-agnostic.

We therefore need a sampling signal that does not depend on a fixed, input-time lookup. The channel CLS tokens (Section~\ref{sec:method:cls}) naturally fill this role: as they process data, they accumulate channel-specific representations. To establish a stable, dataset-level metric of channel relationships, we maintain an exponential moving average (EMA) matrix $S \in \mathbb{R}^{C \times C}$ computed from the channel CLS tokens at a designated backbone layer $\ell^{\star}$:
\begin{equation}
S \leftarrow m\, S + (1 - m)\, \mathrm{sim}\!\left(\mathrm{CLS}^{(\ell^{\star})}\right),
\label{eq:ema}
\end{equation}
where $\mathrm{sim}(\cdot)$ denotes the pairwise cosine similarity function, and $m$ is a momentum close to $1$ so that $S$ reflects stable relations across many batches. The update happens inside the forward method and is decoupled from any downstream loss.

During training, we draw a subset size $C' \sim \mathrm{Uniform}\{1, \ldots, C\}$, and use the relation matrix $S$, scaled by a temperature $T$, to probabilistically guide the selection of the $C'$ channels, biasing the sampler away from purely uniform draws. This allows us to retain the sampling benefits of representation-based subset sampling while remaining channel-agnostic. Channel sampling is disabled at inference.

%% file: 4_experiments.tex
\section{Experiments}
\label{sec:exp}
We evaluate ChanSFormer in supervised classification and self-supervised pretraining regimes on two multi-channel cell benchmarks, CHAMMI, JUMP-CP and a multi-channel satellite benchmark, So2Sat. Because our decomposed spatial-channel attention introduces additional weights in channel attention, we ensure a fair comparison against all small-scale baselines by reducing our backbone's embedding dimension from standard $384$ to $320$ and the number of attention heads to $5$. This adjustment keeps ChanSFormer's total parameter count slightly below the baseline models. 

\subsection{Datasets and metrics}
\label{sec:exp:datasets}

\textbf{CHAMMI}~\cite{chen2023chammi_depthvit} is a multichannel microscopy benchmark comprising roughly 220K images drawn from three sources: the Allen/WTC-11 cell dataset (3 channels), the Human Protein Atlas (HPA, 4 channels), and Cell Painting (CP, 5 channels). To evaluate domain generalization, we follow the established protocol~\cite{pham2024enhancing_dichavit} by extracting feature representations and using a 1-Nearest Neighbour classifier to compute the macro-average F1-score for each task. Following \cite{pham2024enhancing_dichavit}, we report the \emph{averaged score} of the metric from the WTC and HPA subsets.

\textbf{JUMP-CP}~\cite{chandrasekaran2024three_jumpcp} features single-cell images subjected to diverse chemical and genetic perturbations. Our evaluation isolates the compound perturbation plate BR00116991~\cite{bao2023contextual_jumpcpdata}, encompassing 161 classes (160 specific treatments plus one control). Each image contains $8$ channels in total: $5$ capturing fluorescence and $3$ capturing brightfield information. Following \cite{bao2023channel_cvit, pham2024enhancing_dichavit}, we use the official train/val/test split and report the top-1 classification accuracy both on the full $8$-channel input and on the $5$-channel fluorescence-only subset.

\textbf{So2Sat}~\cite{zhu2020so2sat} consists of $32 \times 32$ satellite image patches categorized into 17 distinct local climate zones. The data spans 18 channels per patch, combining 8 Sentinel-1 synthetic aperture radar (SAR) channels and 10 Sentinel-2 multispectral channels. Adopting the city-split partition, our setup uses 352K images for training, 24K for validation and 24K for testing. We report the top-1 classification accuracy under two settings~\cite{pham2024enhancing_dichavit}: \emph{Full} (all 18 channels) and \emph{Sen-1} (the 8 Sentinel-1 radar channels)\CM{.}

\subsection{Implementation details}
\label{sec:exp:impl}

For supervised tasks, we follow the training recipe of DiChaViT~\cite{pham2024enhancing_dichavit}. For self-supervised pretraining, we utilize DINOv2~\cite{oquab2023dinov2} combined with the iBOT~\cite{zhou2021ibot} objective, where we do not use KoLeo regularizer and we use a low weight (0.6) on iBOT loss to stabilize training on small-sized models. We use PyTorch and train all models from scratch. Model training is conducted on distinct hardware setups based on dataset scale: JUMP-CP on $4\times$ RTX PRO 6000 GPUs, So2Sat on $4\times$ A100 GPUs, and CHAMMI on a single Quadro RTX 8000 GPU. Throughput is evaluated on a Quadro RTX 8000. Detailed hyper-parameters are listed in the Appendix.

\subsection{Results}
\label{sec:exp:result}

\subsubsection{Supervised tasks}
\label{sec:exp:supervised}

\begin{table}[t]
\caption{\textbf{Supervised classification on three multichannel benchmarks.} For JUMP-CP and So2Sat, \emph{Full} evaluates on all channels, \emph{Flu.} evaluates on the fluorescence channels for JUMP-CP and \emph{Sen-1} evaluates on the Sentinel-1 channels for So2Sat. The highest number is bold, the second highest is underlined. \#Params is the number of parameters. The reported numbers are the mean of 3 runs.}
\label{tab:supervised_main}
\centering
\small
\setlength{\tabcolsep}{4pt}
\begin{tabular}{lcccccc}
\toprule
 &  & CHAMMI & \multicolumn{2}{c}{JUMP-CP} & \multicolumn{2}{c}{So2Sat} \\
\cmidrule(lr){3-3} \cmidrule(lr){4-5} \cmidrule(lr){6-7}
Method&\#Params & Avg & Full & Flu. & Full & Sen-1 \\
\midrule
HyperViT~\cite{haHypernetworks2016}         & 24.7M & 54.54 & 47.07 & 42.43 & 60.73 & 41.88 \\
DepthwiseViT~\cite{chen2023chammi_depthvit} & 22.6M & 60.94 & 49.86 & 44.98 & 60.41 & 43.41 \\
TemplateViT~\cite{plummerNPAS2022,savarese2018learning} & 21.5M & 57.02 & 52.48 & 43.85 & 55.86 & 37.28 \\
ChAda-ViT~\cite{bourriez2024chada_chadavit} & 21.5M   & 63.88 & 65.03 & 42.15 & 56.98 & 12.38 \\
ChannelViT~\cite{bao2023channel_cvit}       & 21.5M   & 64.90 & 67.51 & 56.49 & 62.75 & 47.70 \\
DiChaViT~\cite{pham2024enhancing_dichavit}  & 21.5M & \textbf{69.68} & \underline{69.19} & \underline{57.98} & \underline{63.36} & \underline{47.76} \\
\midrule
\textbf{ChanSFormer} & 20.1M & \underline{68.83} & \textbf{73.46} & \textbf{61.91} & \textbf{64.76} & \textbf{49.75} \\
\bottomrule
\end{tabular}
\end{table}

We first evaluate ChanSFormer on supervised classification tasks across the three datasets. We compare our approach against a comprehensive suite of channel-adaptive baselines: HyperViT~\cite{haHypernetworks2016}, DepthwiseViT~\cite{chen2023chammi_depthvit}, TemplateViT~\cite{plummerNPAS2022,savarese2018learning}, ChAda-ViT~\cite{bourriez2024chada_chadavit}, ChannelViT~\cite{bao2023channel_cvit}, and DiChaViT~\cite{pham2024enhancing_dichavit}. As detailed in Table~\ref{tab:supervised_main}, ChanSFormer establishes a new SOTA on both JUMP-CP and So2Sat. On JUMP-CP, our model surpasses DiChaViT by $3.93\%$ to $4.27\%$. It demonstrates similarly consistent improvements on So2Sat ($1.40\%$ to $1.99\%$). Notably, these performance gains are achieved using a more compact backbone ($20.1\mathrm{M}$ parameters compared to the $21.5\mathrm{M}$ of recent ViT baselines), proving our architectural efficiency. 
\CM{
On the CHAMMI benchmark, we perform marginally behind DiChaViT by 0.85\%. This is expected as we omit its embedding-dependent Channel Diversification Loss, without which DiChaViT's score drops to 68.07\% \cite{pham2024enhancing_dichavit}, falling below ChanSFormer's 68.83\%.
}
This minor performance trade-off is well worth the architectural flexibility gained for seamlessly handling varying channel configurations.

\subsubsection{Self-supervised tasks}
\label{sec:exp:ssl}

Besides its superior performance in supervised tasks, we evaluate ChanSFormer's representations in a self-supervised regime. We compare DINOv2 pretrained our approach against two baselines: BoC~\cite{de2025scaling}, which achieves channel-agnosticism by processing channels independently, and ChannelViT~\cite{bao2023channel_cvit}, which models cross-channel interactions but relies on channel embeddings. So2Sat is excluded from the pretraining because of its small resolution. To allow dynamic adaptation to varying number of channels, we randomly sample 2 channels for each image during training.
As shown in Table~\ref{tab:ssl}, ChanSFormer outperforms both baselines in nearly all settings. For in-domain linear probing, our model surpasses BoC by $1.32\%$ on CHAMMI and $2.91\%$ on JUMP-CP. The advantage largely persists during cross-dataset transfer, where ChanSFormer leads BoC by up to $2.72\%$. On the other hand, ChannelViT suffers a performance degradation during cross-dataset transfer, lagging behind ChanSFormer by up to $16.88\%$. This collapse occurs because ChannelViT's learned representations are tied to its pretraining channel configuration, forcing it to zero-pad the embeddings for unseen channels in the target dataset. This gap is further exacerbated by a train-eval mismatch, as we sample only two channels in training but use full channels in evaluation. These phenomena demonstrate ChanSFormer's strong generalization to novel channel sets because of its channel-agnostic design.

\begin{table}[h]
\caption{\textbf{Self-supervised linear-probe transfer} across multichannel datasets. Models are pretrained with DINOv2~\cite{oquab2023dinov2} on the dataset listed in the \emph{Pretrain} column, then evaluated by linear probe on the target dataset. \emph{Flu.} and \emph{Sen-1} represent the fluorescence and Sentinel-1 channels respectively. For ChannelViT, the channel embeddings for unseen channels are zero-padded. \protect\CM{$^\dagger$ marks evaluations involving channels absent from SSL pretraining.}
}
\label{tab:ssl}
\centering
\small
\setlength{\tabcolsep}{4pt}
\begin{tabular}{llccccc}
\toprule
 & & CHAMMI & \multicolumn{2}{c}{JUMP-CP} & \multicolumn{2}{c}{So2Sat} \\
 \cmidrule(lr){3-3} \cmidrule(lr){4-5} \cmidrule(lr){6-7}
Pretrain & Method & Avg & Full & Flu. & Full & Sen-1 \\
\midrule
\multirow{3}{*}{CHAMMI}
  & BoC~\cite{de2025scaling}                & 48.92 & 39.36\CRdag & 36.42\CRdag & 67.50\CRdag &  \textbf{46.46}\CRdag \\
  & ChannelViT~\cite{bao2023channel_cvit}   & 44.29 & 23.94\CRdag & 24.71\CRdag & 66.65\CRdag & 44.54\CRdag \\
  & \textbf{ChanSFormer}                  & \textbf{50.24} & \textbf{40.82}\CRdag & \textbf{39.14}\CRdag & \textbf{68.08}\CRdag & {46.23}\CRdag\\
\midrule
\multirow{3}{*}{JUMP-CP \textbf{Flu.}}
  & BoC~\cite{de2025scaling}                & 40.85\CRdag & 46.00\CRdag & 42.24 & 65.47\CRdag & 45.28\CRdag \\
  & ChannelViT~\cite{bao2023channel_cvit}   & 32.88\CRdag & 30.72\CRdag & 32.97 & 65.00\CRdag & 44.47\CRdag \\
  & \textbf{ChanSFormer}                  & \textbf{42.86}\CRdag & \textbf{46.13}\CRdag & \textbf{45.15} & \textbf{66.33}\CRdag & \textbf{45.34}\CRdag \\
\bottomrule
\end{tabular}
\end{table}

\begin{table}[t]
\centering
\begin{minipage}[t]{0.44\linewidth}
\caption{\textbf{Model-level throughput (images/sec).} Higher is better. Benchmarked on 8 channel 224$\times$224 inputs. \protect\CM{ChanSFormer speeds up both training and inference over global attention. Naive is naive attention as in Table~\ref{tab:speed_kernel}.}}
\label{tab:speed_e2e}
\centering
\small
\setlength{\tabcolsep}{2pt}
\resizebox{\linewidth}{!}{%
\begin{tabular}{l l c c}
\toprule
Mode & Model & Patch $16$ & Patch $8$ \\
\midrule
\multirow{3}{*}{Inference} & ChannelViT & $72.2$ & $\phantom{0}7.7$ \\
                           & \CM{ChanSFormer (naive)} & \CM{$106.7$} & \CM{$22.0$} \\
                           & \textbf{ChanSFormer} & $\mathbf{112.0}$ & $\mathbf{22.9}$ \\
\midrule
\multirow{3}{*}{Training}  & ChannelViT & $16.3$ & $\phantom{0}1.5$ \\
                           & \CM{ChanSFormer (naive)} & \CM{$30.8$} & \CM{$\phantom{0}5.8$} \\
                           & \textbf{ChanSFormer} & $\mathbf{33.9}$ & $\phantom{0}\mathbf{6.2}$ \\
\bottomrule
\end{tabular}%
}
\end{minipage}
\hfill
\begin{minipage}[t]{0.54\linewidth}
\caption{\textbf{Speed and GPU memory for channel attention.} Benchmarked on input tensors of shape $(B, C, N+1, D) = (64, 8, 197, 384)$. Lower is better. Our optimized attention reduces execution time and GPU memory footprint compared to standard implementations.}
\label{tab:speed_kernel}
\centering
\small
\setlength{\tabcolsep}{3pt}
\begin{tabular}{lcccc}
\toprule
 & \multicolumn{2}{c}{Inference} & \multicolumn{2}{c}{Training} \\
\cmidrule(lr){2-3} \cmidrule(lr){4-5}
Method & Speed & Memory & Speed & Memory \\
\midrule
Naive attention  & $7.1$\,ms & $1029$\,MB & $26.3$\,ms & $1117$\,MB \\
Mem-efficient    & $6.0$\,ms & $\phantom{0}797$\,MB  & $18.6$\,ms & $2283$\,MB \\
\textbf{Ours} & $\mathbf{4.9}$\,ms & $\phantom{0}\mathbf{634}$\,MB & $\mathbf{12.1}$\,ms & $\phantom{0}\mathbf{877}$\,MB \\
\bottomrule
\end{tabular}
\end{minipage}
\end{table}

\subsubsection{Computational efficiency}
\label{sec:exp:speed}

\textbf{Model-level throughput.}
Table~\ref{tab:speed_e2e} demonstrates the throughput improvement yielded by ChanSFormer's overall architecture. By disentangling the $O((CN)^2)$ global attention into sequential channel and spatial attention with $O(CN^2 + NC^2)$ complexity, ChanSFormer alleviates the primary computational bottleneck. \CM{This reduction alone, measured with naive channel attention, accelerates inference by $48\%$--$186\%$ and training by $89\%$--$286\%$ over ChannelViT. Our optimized kernel, which amortizes the fixed overheads of short channel sequences, adds a further $4\%$--$10\%$ over the naive implementation. Together, they increase the inference throughput of the full model by $55\%$ at patch size 16 and $197\%$ at patch size 8, and the training throughput by $108\%$ to $313\%$.} The performance advantage widens drastically as the spatial sequence length $N$ grows, matching the predicted scaling for multichannel processing.

\textbf{Hardware-aware channel attention.}
Table~\ref{tab:speed_kernel} isolates the performance of our optimized channel attention kernel with an input of 8 channels ($C = 8$). Standard hardware-aware implementations, such as memory-efficient attention \cite{xFormers2022}, are heavily optimized for long sequences. They struggle with our extremely short channel sequence regime, which inflates their training workspace memory to $2283\,\mathrm{MB}$ due to the fixed overheads discussed in Section \ref{sec:method:attn}. Our optimized kernel sidesteps these issues by fusing operations and adapting to small tile sizes. It reduces the running time during the full training step by ${35\%}$ to ${54\%}$, while simultaneously reducing the memory footprint by 20\% to 62\%.

\subsubsection{\texorpdfstring{\CM{Robustness to missing and novel channels}}{Robustness to missing and novel channels}}
\label{sec:exp:missing_channels}

Following \cite{bao2023channel_cvit, pham2024enhancing_dichavit}, we further evaluate the model's robustness by testing its supervised classification performance on partial channels of the JUMP-CP dataset. We systematically drop channels and report the average accuracy across all $\binom{8}{k}$ combinations of $k$ available channels. As shown in Table~\ref{tab:jumpcp_partial}, ChanSFormer \CM{outperforms the baselines whenever at least two channels are available, leading DiChaViT by 0.20\% to 3.70\% with 2--7 channels}, highlighting the robustness of our channel-agnostic design \CM{to missing channels. This setting removes subsets of channels that were observed during training, and therefore measures robustness to missing channels rather than generalization to channels with unseen semantics.}

\begin{table}[h]
\caption{\textbf{Test accuracy under partial channels of JUMP-CP.} Each column reports the mean $\pm$ std over all $\binom{8}{k}$ combinations of $k$ available channels out of $8$~\cite{bao2023channel_cvit,pham2024enhancing_dichavit}. ChanSFormer achieves comparable or better performance in all 8 cases.}
\label{tab:jumpcp_partial}
\centering
\small
\setlength{\tabcolsep}{2pt}
\resizebox{\linewidth}{!}{%
\begin{tabular}{lcccccccc}
\toprule
 & \multicolumn{8}{c}{Number of channels for evaluation ($k$)} \\
\cmidrule(lr){2-9}
Method & $8$ & $7$ & $6$ & $5$ & $4$ & $3$ & $2$ & $1$ \\
\midrule
ChannelViT~\cite{bao2023channel_cvit} & 67.51 & $60.36_{\pm 9.1}$ & $52.74_{\pm 12.2}$ & $44.89_{\pm 13.2}$ & $36.88_{\pm 12.3}$ & $29.36_{\pm 9.3}$ & $23.70_{\pm 5.0}$ & $20.78_{\pm 1.6}$ \\
DiChaViT~\cite{pham2024enhancing_dichavit} & 69.19 & $61.91_{\pm 9.3}$ & $54.49_{\pm 12.4}$ & $46.35_{\pm 13.4}$ & $38.00_{\pm 12.4}$ & $30.09_{\pm 9.3}$ & $23.97_{\pm 4.9}$ & $20.90_{\pm 1.6}$ \\
\midrule
\textbf{ChanSFormer} & \textbf{73.46} & $\bm{65.61}_{\pm 10.0}$ & $\bm{57.32}_{\pm 12.8}$ & $\bm{48.52}_{\pm 13.7}$ & $\bm{39.50}_{\pm 12.5}$ & $\bm{30.92}_{\pm 9.2}$ & $\bm{24.17}_{\pm 4.6}$ & $20.65_{\pm 1.3}$ \\
\bottomrule
\end{tabular}%
}
\end{table}

\CM{We next evaluate the generalization of ChanSFormer to novel channels in the self-supervised setting. Unlike the partial-channel setting above, which only removes channels observed during training, the cross-dataset transfer in Table~\ref{tab:ssl}, following CA-MAE~\cite{kraus2024masked_camae} and BoC~\cite{de2025scaling}, already evaluates transfer to channels whose semantics are absent from pretraining (marked by $^\dagger$). In addition, we conduct an experiment on CHAMMI in which either HPA or Allen is excluded from SSL pretraining, using the same pretraining recipe as in Section~\ref{sec:exp:ssl}. Every evaluation in this setting therefore involves channels that are absent from pretraining. As shown in Table~\ref{tab:ssl_loso}, ChanSFormer outperforms both baselines in five of the six settings and remains far ahead of ChannelViT, whose zero-padded embeddings for unseen channels leave it near $19.5\%$ on JUMP-CP. Thus, the advantage of ChanSFormer extends beyond robustness to missing channels to the novel-channel transfer regime that motivates this work.}

\begin{table}[t]
\CMcolor
\caption{\textbf{Leave-one-source-out self-supervised transfer.} Models are pretrained with DINOv2~\cite{oquab2023dinov2} on CHAMMI with one source held out, then evaluated by linear probe (JUMP-CP) or the official CHAMMI protocol. All evaluations in this table involve channels absent from SSL pretraining. Bold denotes the best result under each pretraining source.}
\label{tab:ssl_loso}
\centering
\small
\setlength{\tabcolsep}{4pt}
\begin{tabular}{llccc}
\toprule
 & & CHAMMI & \multicolumn{2}{c}{JUMP-CP} \\
 \cmidrule(lr){3-3} \cmidrule(lr){4-5}
Pretrain & Method & Avg & Full & Flu. \\
\midrule
\multirow{3}{*}{\shortstack[l]{CHAMMI: Allen + CP\\(HPA held out)}}
  & BoC~\cite{de2025scaling}                & 35.72 & 19.63 & 19.65 \\
  & ChannelViT~\cite{bao2023channel_cvit}   & 33.92 & 19.54 & 19.53 \\
  & \textbf{ChanSFormer}                    & \textbf{46.00} & \textbf{38.38} & \textbf{37.25} \\
\midrule
\multirow{3}{*}{\shortstack[l]{CHAMMI: HPA + CP\\(Allen held out)}}
  & BoC~\cite{de2025scaling}                & \textbf{46.76} & 37.75 & 35.67 \\
  & ChannelViT~\cite{bao2023channel_cvit}   & 40.93 & 19.62 & 19.62 \\
  & \textbf{ChanSFormer}                    & 44.38 & \textbf{38.58} & \textbf{37.58} \\
\bottomrule
\end{tabular}
\end{table}

\subsubsection{Ablation studies}
\label{sec:exp:ablations}

\begin{table}[t]
\caption{\textbf{Ablation on channel embeddings and cross-channel interactions.} Channel embeddings are not critical for ChanSFormer compared with ChannelViT with global attention, and cross‑channel interactions prove essential. \emph{Flu.} and \emph{Sen-1} represents the fluorescence and Sentinel-1 channels respectively. }
\label{tab:ablation_channel_attn}
\centering
\small
\setlength{\tabcolsep}{4pt}
\begin{tabular}{lccccc}
\toprule
 & CHAMMI & \multicolumn{2}{c}{JUMP-CP} & \multicolumn{2}{c}{So2Sat} \\
 \cmidrule(lr){2-2} \cmidrule(lr){3-4} \cmidrule(lr){5-6}
Method & Avg & Full & Flu. & Full & Sen-1 \\
\midrule
\textbf{ChanSFormer} & 68.83 & 73.46 & 61.91 & 64.76 & 49.75 \\
\quad w. channel embedding & 69.03 & 73.78 & 62.36 & 64.78 & 49.55 \\
\midrule
ChannelViT~\cite{bao2023channel_cvit} & 64.90 & 67.51 & 56.49 & 62.75 & 47.70\\
\quad w/o channel embedding (CA-MAE~\cite{kraus2024masked_camae}) & 50.37 & 61.93 & 51.56 & 62.58 & 46.73\\
\midrule
Spatial-only (BoC~\cite{de2025scaling} + CLS channel attn) & 61.73 & 41.56 & 39.10 & 63.41 & 47.70 \\
\bottomrule
\end{tabular}
\end{table}

\textbf{Ablation on channel embeddings.}
Table~\ref{tab:ablation_channel_attn} shows the reliance of different architectures to fixed channel embeddings. Adding channel embeddings to ChanSFormer yields only marginal improvements. In stark contrast, removing them from ChannelViT causes a catastrophic performance degradation (e.g., a $14.53$ drop on CHAMMI). Note that this embedding-free ChannelViT represents the architectural backbone of CA-MAE~\cite{kraus2024masked_camae}, an existing channel-agnostic method. These results demonstrate that our decomposed attention and individual CLS tokens natively preserve channel identities, rendering hard-coded embeddings obsolete.

\textbf{Ablation on cross-channel interactions.}
We further ablate the necessity of modeling cross-channel correlations. The ``Spatial-only'' baseline in Table~\ref{tab:ablation_channel_attn} processes spatial patches independently, effectively adapting the channel-agnostic approach of BoC~\cite{de2025scaling} to supervised tasks by utilizing the same query-based attention pooling described in Section~\ref{sec:method:arch}. The severe performance degradation observed across all datasets confirms that late-stage feature aggregation is insufficient for complex multi-channel data. These results demonstrate that cross-channel correlations are vital for robust modeling of multi-channel imagery.

\textbf{Ablation on channel sampling and CLS tokens.}
The first three rows of Table~\ref{tab:ablation_design} sequentially ablate our architectural components in the supervised setting. First, replacing our representation-based sampling with purely uniform sampling (HCS \cite{bao2023channel_cvit}) causes a \CM{consistent} performance drop ($2.66\%$ on CHAMMI\CM{, $0.65\%$ on JUMP-CP Full, and $0.13\%$ on JUMP-CP Flu.}). Next, building upon this HCS baseline, we replace the individual channel CLS tokens with a single CLS token \cite{bertasius2021space_timesformer}. In this TimeSformer-style baseline, the CLS token is copied during spatial attention and subsequently averaged across  channels. This modification further degrades accuracy across all datasets. Because individual channels carry highly heterogeneous information, average pooling them into a single token forcibly destroys critical modality-specific features. Therefore, maintaining independent channel tokens is essential for cell painting images.

\textbf{Ablation on SSL feature granularity.}
We further ablate the granularity of the supervisory signals in our self-supervised learning (SSL) framework, comparing channel-independent tokens against an aggregated image-level token. As shown in the lower half of Table~\ref{tab:ablation_design}, our approach consistently achieves superior linear-probe performance, regardless of the pretraining dataset (CHAMMI or JUMP-CP). This advantage stems from the fact that different channels carry highly heterogeneous information. Retaining independent CLS tokens allows the model to utilize this heterogeneity as distinct, modality-specific supervisory signals during pretraining. This provides a richer supervisory signal per sample and increases the diversity of training signals within each batch, thus improving the performance. \CM{This comparison also separates the contribution of the backbone from that of the supervision granularity. With image-level attention pooling, ChanSFormer is trained with the same supervision granularity as ChannelViT, yet it outperforms ChannelViT (Table~\ref{tab:ssl}) in all six settings, mostly by a large margin, indicating that the backbone provides the main improvement. Channel-level supervision then adds a further gain in all six settings.}

\begin{table}[t]
\caption{\textbf{Ablation of ChanSFormer's design choices.} The \emph{Setting} column distinguishes between supervised classification and DINOv2 linear probes. For the DINOv2 ablation, models are pretrained on either CHAMMI or JUMP-CP (indicated as PT) to compare feature granularity. \emph{Flu.} represents the fluorescence channels.}
\label{tab:ablation_design}
\centering
\small
\setlength{\tabcolsep}{2.4pt}
\begin{tabular}{llccc}
\toprule
Setting & Method & CHAMMI & JUMP-CP Full & JUMP-CP Flu. \\
\midrule
\multirow{3}{*}{Supervised} 
& \textbf{ChanSFormer} & \textbf{68.83} & \textbf{73.46} & \textbf{61.91} \\
& \quad w. HCS \cite{bao2023channel_cvit} channel sampling & 66.17 & 72.81 & 61.78 \\
& \quad w. HCS \cite{bao2023channel_cvit} w/o channel CLS token \cite{bertasius2021space_timesformer} & 64.58 & 69.77 & 58.00\\
\midrule
DINOv2 & \textbf{ChanSFormer (channel-level)} & \textbf{50.24} & \textbf{40.82} & \textbf{39.14} \\
(PT: CHAMMI) & \quad w/ attention pooling (image-level) & 48.61 & 38.97 & 38.52 \\
\midrule
DINOv2 & \textbf{ChanSFormer (channel-level)} & \textbf{42.86} & \textbf{46.13} &  \textbf{45.15} \\
(PT: JUMP-CP Flu.) & \quad w/ attention pooling (image-level) & 41.84 & 45.95 & \CM{45.10} \\
\bottomrule
\end{tabular}
\end{table}

%% file: 5_conclusion.tex
\section{Conclusion}
\label{sec:conclusion}

We presented ChanSFormer, a highly efficient, channel-agnostic Vision Transformer that replaces rigid channel embeddings with decomposed spatial-channel attention and individual channel CLS tokens. This design preserves distinct channel identities and cross-channel interactions naturally, while also unlocking superior self-supervised representation learning. Empirically, ChanSFormer achieves comparable or superior performance compared to SOTA methods across two biology multi-channel datasets and one satellite dataset. Furthermore, it demonstrates exceptional cross-dataset transferability and accelerates model throughput. By eliminating the need for channel-specific retraining, our model provides a unified, flexible backbone that paves the way for adaptable foundation models in multi-channel imaging.

\textbf{Limitations.}
While our channel-agnostic design provides architectural flexibility, omitting fixed channel embeddings precludes the \CM{direct use} of auxiliary channel-diversity losses~\cite{pham2024enhancing_dichavit}, leading to \CM{slightly lower performance on CHAMMI}. \CM{Adapting them is nontrivial: they are defined over static embeddings, whereas our channel CLS tokens are data-dependent representations, and directly forcing them apart may suppress meaningful cross-channel correlations. Developing a channel-agnostic diversification objective is a promising direction.} Furthermore, due to resource constraints, 
scaling ChanSFormer to the data and model size of foundation models like BoC~\cite{de2025scaling} is left to future work.

%% file: 6_supp.tex
\newpage
\appendix

\section{Details of the channel attention kernel}
\label{supp:channel_attn_kernel}

We provide here the design and implementation details of the channel attention kernel. The kernel is written in Triton and supports both the forward and backward passes, with full autograd integration. Our channel size $C$ ranges from 1 to 18 in practice, which sits well below the granularity at which mainstream attention kernels are tuned. We organize the discussion around three design choices: (i) a $C$-adaptive dispatch between two kernel paths, (ii) instance batching to amortize memory traffic, and (iii) a stride-aware launch interface that eliminates layout-conversion copies.

\paragraph{$C$-adaptive dispatch.}
NVIDIA TensorCores require the inner dimensions of a matrix multiplication to be at least 16 in order to issue a single MMA instruction; below this threshold, the operands must be padded, and the padded lanes contribute no useful work. For channel attention with $C < 16$, the score matrix $S = QK^\top \in \mathbb{R}^{C \times C}$ and the output matmul $PV$ both fall into this regime, so a naive use of TensorCores wastes most of the issued FLOPs on padding. We therefore dispatch according to the length of channel sequences $C$ at kernel-launch time. For short channel sequences ($C\leq8$), we avoid using TensorCores entirely and instead use the general-purpose CUDA cores, which have no minimum-shape requirement and are not bottlenecked at this scale. For longer sequences ($C>8$), we revert to the standard TensorCores as they are faster at this scale. The crossover between paths is determined empirically on our target hardware\CM{, an NVIDIA Quadro RTX 8000: we benchmark both paths across the practical channel-count range and select the point at which the TensorCore path becomes faster. On this GPU,} the small path remains faster up through $C=8$, after which the large path takes over. \CM{Because both the padding overhead and the TensorCore throughput depend on the GPU architecture, this threshold is hardware-dependent. On other devices, it can be selected by a lightweight device-specific microbenchmark at installation or initialization time.}

\paragraph{Instance batching for short-sequence amortization.}
A single channel attention call in our model spawns hundreds to thousands of independent attention instances (one per spatial position, per head, across the batch). If each instance were launched as a separate GPU program, as in a direct port of FlashAttention~\cite{dao2022flashattention}, the per-instance computation would be far too small to amortize the fixed costs of kernel launch, register allocation, and shared memory setup. To eliminate this overhead on the small-$C$ path, we let each program process a batch of instances together, with the batch size automatically tuned for each $(C, \text{HEAD\_DIM})$ configuration. All operations---loading queries, keys, and values, the softmax reduction, and writing the result---are then executed in a batched manner, so that the launch cost is roughly divided by the batch size. When $C$ becomes large, a single instance already provides enough parallelism to fully occupy the GPU's compute units, and batching is no longer needed; we fall back to processing one instance per program. We further prune any configuration that our cost model predicts will exceed register capacity and cause spilling.

\paragraph{Stride-aware launch interface.}
A subtle but important source of overhead in channel attention stems from layout conversion. Most attention kernels require the attention axis to occupy a fixed tensor dimension, which in our case would force an explicit permutation and a contiguous copy on every call, together with a symmetric copy when restoring the original shape. At the batch sizes we use, these copies already cost on par with the attention computation itself. We remove this overhead by making the kernel stride-aware. The launch wrapper inspects the input strides and passes the necessary stride metadata to the kernel. The kernel then uses that stride information to compute load and store offsets, handling both standard layouts and the native multidimensional layout of our queries, keys, and values with zero data movement.

\paragraph{Backward pass.}
The backward pass mirrors the forward dispatch. Following the standard FlashAttention recurrence, we save only the log-sum-exp statistics from the forward pass and recompute $S$, $P$, and $O$ inside the backward kernel; this trades a modest amount of additional FLOPs for a substantial reduction in saved-activation memory. The three gradient tensors $dQ$, $dK$, $dV$ are computed in a single fused kernel per path.

\FloatBarrier
\section{Extended implementation details}
\label{sec:supp:hparams}

\paragraph{Supervised Training Hyperparameters.}
For supervised classification tasks, we optimize all models using AdamW. For JUMP-CP and So2Sat, we apply a cosine learning rate schedule with a 10 epoch linear warmup, a base learning rate of $5 \times 10^{-4}$, and a weight decay of $0.04$. They are trained for $100$ epochs using standard cross-entropy loss, with batch sizes of $256$ (patch size $16$) and $128$ (patch size $8$), respectively. For CHAMMI, we use a cosine learning rate schedule with a 3 epoch linear warmup, a learning rate of $4 \times 10^{-5}$ and train for $60$ epochs with a batch size of $64$, a patch size of $16$ and proxy loss. For the representation-based channel sampling applied across these experiments, we use an EMA momentum of $m = 0.9995$, a scaling temperature of $T = 0.01$ for CHAMMI. We use an EMA momentum of $m = 0.9995$, a scaling temperature of $T = 0.1$ for So2Sat. We use an EMA momentum of $m = 0.998$, a scaling temperature of $T = 1.0$ for JUMP-CP. \CM{Following DiChaViT~\cite{pham2024enhancing_dichavit}, $T$ is selected separately for each dataset on its validation set; its sensitivity is analyzed in Table~\ref{tab:temp_sens}.}

\paragraph{Representation-based channel sampling.}
Varying the input channel set during training serves as an effective regularizer for our channel-agnostic backbone. The prior method DiChaViT \cite{pham2024enhancing_dichavit} samples a random channel and then sample additional ones based on their dissimilarity to that initial selection. However, we observe that this heuristic often leads to an unintended outcome: selected channels may be distant from the initial anchor while remaining highly similar to each other. Our empirical analysis of this process confirms that the vast majority of channel pairs selected by this approach are indeed  similar pairs. Consequently, in our representation-based sampling, we intentionally and directly sample subsets of similar channels. As demonstrated in Table \ref{tab:ablation_design}, this strategy effectively regularizes the training process and yields superior performance. Specifically, we compute channel similarities at layer $\ell^{\star}=6$. We find that representations from initial or very late layers yield slightly inferior performance, while the middle-to-late range (layers 6--9) provides the better signal for representation-based sampling.
\CM{Moreover, the diagonal entries of $S$ in Eq.~\eqref{eq:ema}, which represent self-similarity, are excluded from sampling. On CHAMMI, whose three sources contribute disjoint channel sets (Allen: 3 channels, HPA: 4, CP: 5), we follow the channel indexing of DiChaViT~\cite{pham2024enhancing_dichavit} and assign the channels distinct global indices, giving $C = 12$. Although $S$ is stored as a single $12 \times 12$ buffer, it is updated and queried only within the block that corresponds to the source of the current image. $S$ therefore effectively consists of three independent $3 \times 3$, $4 \times 4$, and $5 \times 5$ similarity matrices, with an effective $C$ of 3, 4, or 5 depending on the source; cross-source entries are never updated or used for sampling.} \CM{$S$ is a non-trainable buffer: the channel CLS tokens are detached before the similarities are computed, so no gradient flows through $S$ or the sampling operation. In distributed training, the batch similarity matrix is averaged across all workers before the EMA update, so that every worker maintains the same $S$; since $S$ is at most $18 \times 18$, the synchronization overhead is negligible.}

\paragraph{Self-Supervised Pretraining Hyperparameters.}
In our self-supervised framework based on DINOv2~\cite{oquab2023dinov2} with iBOT~\cite{zhou2021ibot}, we adapt the default loss weighting to accommodate the presence of multiple independent channel CLS tokens. Specifically, the KoLeo regularizer weight is removed to prevent NaN and improve performance. The iBOT loss weight is reduced to 0.6 to ensure stable training. \CM{We sample only 2 channels for each image. This allows dynamic adaptation to varying numbers of channels, enlarges the batch size under a fixed memory budget, and provides a unified protocol across datasets whose images contain different numbers of channels (e.g., 3, 4, or 5 on CHAMMI).} The batch size is \CM{set to} 1024 for BoC and 512 for ChannelViT and ChanSFormer, \CM{so that the three models see the same number of channels per iteration}. We use a schedule of $100$ channel-epochs on CHAMMI and JUMP-CP. The channel-epoch refers to \CM{the number of} iterations that allows the model to see the total number of channels in the dataset (\CM{e.g., assuming} every sample has 4 channels, $100$ channel-epochs is 200 epochs as we sample 2 channels for each sample). The base learning rate is set to 0.002 with a batch size of 1024 and the same square root scaling rule for learning rate as the \CM{vanilla} DINOv2. \CM{Although each pretraining view contains only two channels, the frozen backbone is evaluated with up to 8 channels on JUMP-CP and 18 channels on So2Sat (Table~\ref{tab:ssl}). ChanSFormer imposes no maximum channel count, because its patch embedding, spatial positional embedding, and attention operations are all shared across the channel axis, so it operates on substantially more channels than it has observed simultaneously during pretraining.}

For linear-probe evaluations on the JUMP-CP and So2Sat dataset, we use concatenated channel cls tokens from the last 4 layers and follow  the same protocol in the vanilla DINOv2. For ChannelViT where we do not have the channel level cls tokens, we use the averaged-pooled token of each channel along with the single global cls token. On the CHAMMI dataset, we also concatenate channel level cls tokens but only use the features from the last layer and we follow the standard evaluation protocol as \cite{pham2024enhancing_dichavit}.

\FloatBarrier
\section{\texorpdfstring{Standard \CM{deviation} of the supervised results}{Standard deviation of the supervised results}}
\label{sec:supp:std}
Here we reported the standard deviations of our method in Table \ref{tab:supp:supervised_std}.

\begin{table}[!htb]
\centering
\small
\caption{Standard deviation of ChanSFormer on 3 datasets.}
\label{tab:supp:supervised_std}
\setlength{\tabcolsep}{6pt}
\begin{tabular}{lccccc}
\toprule
Method & CHAMMI & JUMP-CP Full & JUMP-CP Flu. & So2Sat Full & So2Sat Sen-1 \\
\midrule
\textbf{ChanSFormer} & 68.83 $\pm$ 0.51 & 73.46 $\pm$ 0.14 & 61.91 $\pm$ 0.62 & 64.76 $\pm$ 0.30 & 49.75 $\pm$ 0.33 \\
\bottomrule
\end{tabular}
\end{table}

\FloatBarrier
\section{Subset results on the CHAMMI dataset}
\label{sec:supp:chammi_subset}
We detail the subset evaluation on the CHAMMI dataset, as shown in Table \ref{tab:chammi_subsets_detailed} and Table \ref{tab:ssl_chammi_subsets_final}. In supervised learning, we surpass ChannelViT in average accuracy. We achieved a bit lower performance on the Allen and CP subsets compared with DiChaViT, which we attribute to the inability to leverage auxiliary channel-diversity losses when fixed channel embeddings are omitted. In the DINOv2 linear-probe setting, our approach demonstrates superior performance on the HPA subset and yields the highest overall accuracy across both in-domain and cross-dataset evaluations.

\begin{table}[!htb]
\centering
\small
\caption{Detailed breakdown of performance on CHAMMI subsets (Allen, HPA, and CP). The Avg column represents the mean of Allen and HPA. Baseline results are reported from DiChaViT~\cite{pham2024enhancing_dichavit}.}
\label{tab:chammi_subsets_detailed}
\setlength{\tabcolsep}{4pt}
\begin{tabular}{lccccc}
\toprule
Method & \#Params & Allen & HPA & CP & Avg \\
\midrule
HyperViT~\cite{haHypernetworks2016} & 24.7M & 45.17 & 63.90 & 26.23 & 54.54 \\
DepthwiseViT~\cite{chen2023chammi_depthvit} & 22.6M & 50.35 & 71.52 & 27.74 & 60.94 \\
TemplateViT~\cite{plummerNPAS2022} & 21.5M & 49.51 & 64.52 & 25.65 & 57.02 \\
ChAda-ViT~\cite{bourriez2024chada_chadavit} & 21.5M & 67.08 & 60.67 & 24.60 & 63.88 \\
ChannelViT~\cite{bao2023channel_cvit} & 21.5M & 67.66 & 62.14 & 27.62 & 64.90 \\
DiChaViT~\cite{pham2024enhancing_dichavit} & 21.5M & 75.69 & 63.67 & 28.98 & 69.68 \\
\midrule
\textbf{ChanSFormer} & 20.1M & 69.00 & 68.67 & 27.31 & 68.83 \\
\bottomrule
\end{tabular}
\end{table}

\begin{table}[!htb]
\centering
\small
\caption{Detailed breakdown of performance on CHAMMI subsets (Allen, HPA, and CP) for DINOv2 linear-probe. The Avg column represents the mean of Allen and HPA.}
\label{tab:ssl_chammi_subsets_final}
\setlength{\tabcolsep}{3pt}
\begin{tabular}{llcccc}
\toprule
Pretrain & Method & Allen & HPA & CP & Avg \\
\midrule
\multirow{3}{*}{CHAMMI} 
& BoC~\cite{de2025scaling} & \textbf{49.88} & 47.96 & \textbf{27.91} & 48.92 \\
& ChannelViT~\cite{bao2023channel_cvit} & 46.60 & 41.98 & 26.93 & 44.29 \\
& \textbf{ChanSFormer} & 48.96 & \textbf{51.51} & 25.10 & \textbf{50.24} \\
\midrule
\multirow{3}{*}{JUMP-CP Flu.} 
& BoC~\cite{de2025scaling} & 43.31 & 38.40 & 27.83 & 40.85 \\
& ChannelViT~\cite{bao2023channel_cvit} & 32.48 & 33.29 & \textbf{28.77} & 32.88 \\
& \textbf{ChanSFormer} & \textbf{44.48} & \textbf{41.25} & 27.99 & \textbf{42.86} \\
\bottomrule
\end{tabular}

\end{table}

\FloatBarrier
\section{Supervised results with ImageNet pretraining}
\label{sec:supp:imagenet_pretrained}
The main paper trains all models from scratch (Table~\ref{tab:supervised_main}). Here we additionally report results when models are initialized from an ImageNet pretrained ViT-S/16, which is the more common setting on general benchmarks. We use the same training recipe as Section~\ref{sec:exp:impl} with shorter schedules and lower learning rates suitable for fine-tuning: 45 epochs on CHAMMI (lr $4\times10^{-5}$, 3 epoch warmup, batch size 64), 50 epochs on JUMP-CP (lr $1\times10^{-4}$, 5 epoch warmup, batch size 256), and 20 epochs on So2Sat (lr $1\times10^{-4}$, 2 epoch warmup, batch size 256). We did not include ChAda-ViT as it is ChannelViT without HCS in this case. To make the spatial-attention weights compatible with the ImageNet ViT-S/16 checkpoint, ChanSFormer in this setting uses the standard ViT-S embedding dimension $D = 384$ rather than the $D = 320$ used in the main paper; this raises its parameter count to $28.9\mathrm{M}$. Table~\ref{tab:supp:imagenet_pretrained} summarizes the results on the full channel set of each dataset. ChanSFormer achieves the highest accuracy on all three datasets, surpassing the best baselines by $0.76\%$ to $1.5\%$. The disentangled architecture transfers well from natural-image pretraining and remains the strongest channel-adaptive ViT in this setting without relying on a learnable channel embedding.

\begin{table}[!htb]
\caption{\textbf{Supervised classification with ImageNet ViT-S/16 pretraining.} Top-1 accuracy on the full channel set of each dataset. The highest number per column is bold and the second highest is underlined.}
\label{tab:supp:imagenet_pretrained}
\centering
\small
\setlength{\tabcolsep}{4pt}
\begin{tabular}{lcccc}
\toprule
Method & \#Params & CHAMMI & JUMP-CP & So2Sat \\
\midrule
HyperViT~\cite{haHypernetworks2016} & 24.7M & 62.28 & 39.64 & 61.22 \\
DepthwiseViT~\cite{chen2023chammi_depthvit} & 22.6M & 76.52 & 46.97 & 61.55 \\
TemplateViT~\cite{plummerNPAS2022,savarese2018learning} & 21.5M & 70.30 & 35.01 & 60.84 \\
ChannelViT~\cite{bao2023channel_cvit} & 21.5M & \underline{80.68} & 80.81 & \underline{68.01} \\
DiChaViT~\cite{pham2024enhancing_dichavit} & 21.5M & 78.83 & \underline{82.99} & 67.18 \\
\midrule
\textbf{ChanSFormer} & 28.9M & \textbf{82.18} & \textbf{84.11} & \textbf{68.77} \\
\bottomrule
\end{tabular}
\end{table}

\FloatBarrier
\section{\texorpdfstring{\CM{Sensitivity to the sampling temperature}}{Sensitivity to the sampling temperature}}
\label{sec:supp:temperature}
\CM{The temperature $T$ controls how strongly the learned channel relations shape the sampling distribution: large values approach uniform HCS sampling, whereas small values concentrate the samples on the channel combinations favored by $S$. Table~\ref{tab:temp_sens} varies $T$ around the selected value on each dataset. On So2Sat, $T = 0.1$ and $1.0$ achieve nearly identical overall performance, while $T = 0.01$ is slightly lower, mainly on Sentinel-1. On JUMP-CP, $T = 1.0$ and $10$ perform similarly, while $T = 0.1$ leads to a noticeable drop. On CHAMMI, $T = 0.01$ and $0.1$ are comparable, whereas $T = 0.001$ and $1.0$ perform worse. Performance is therefore stable within an appropriate range for each dataset, while excessively concentrating or flattening the sampling distribution reduces it.
The selected values differ across datasets partly because $S$ has a different contrast on each: the standard deviation of its off-diagonal entries is $0.134$--$0.174$ on the three CHAMMI sources, $0.203$ on So2Sat, and $0.295$ on JUMP-CP. This roughly two-fold difference does not fully account for the two orders of magnitude spanned by the selected $T$, since $T$ also determines how strongly the learned relations are sharpened. Similarly, although JUMP-CP and So2Sat share a coarse two-block structure in $S$ (Figures~\ref{fig:channel_relations} and~\ref{fig:channel_relations_so2sat}), this does not imply the same optimal $T$. The ten-fold difference between their selected values does not indicate sharply different optima, however: $T = 1.0$ performs well on both datasets, nearly matching $T = 0.1$ on So2Sat and giving the best result on JUMP-CP.}

\begin{table}[!htb]
\CMcolor
\caption{\textbf{Sensitivity to the sampling temperature $T$} in supervised training. The selected value for each dataset is shown in bold in the $T$ column; bold numbers denote the best result in each column.}
\label{tab:temp_sens}
\centering
\small
\begin{minipage}[t]{0.36\linewidth}
\centering
\begin{tabular}{lcc}
\toprule
$T$ & So2Sat Full & Sen-1 \\
\midrule
0.01 & 64.85 & 48.35 \\
\textbf{0.1} & 64.76 & \textbf{49.75} \\
1.0 & \textbf{64.99} & 49.46 \\
\bottomrule
\end{tabular}
\end{minipage}
\hfill
\begin{minipage}[t]{0.36\linewidth}
\centering
\begin{tabular}{lcc}
\toprule
$T$ & JUMP-CP Full & Flu. \\
\midrule
0.1 & 71.41 & 58.39 \\
\textbf{1.0} & \textbf{73.46} & \textbf{61.91} \\
10 & 73.12 & 61.18 \\
\bottomrule
\end{tabular}
\end{minipage}
\hfill
\begin{minipage}[t]{0.24\linewidth}
\centering
\begin{tabular}{lc}
\toprule
$T$ & CHAMMI \\
\midrule
0.001 & 67.34 \\
\textbf{0.01} & \textbf{68.83} \\
0.1 & 68.40 \\
1.0 & 66.26 \\
\bottomrule
\end{tabular}
\end{minipage}
\end{table}

\FloatBarrier
\section{\texorpdfstring{\CM{Number of channels per view in self-supervised pretraining}}{Number of channels per view in self-supervised pretraining}}
\label{sec:supp:ssl_channels}
\CM{To examine the effect of using more channels per view, we additionally pretrain on JUMP-CP Flu.\ with four channels per image, reducing the image batch size from 512 to 256 so that the total number of channel views per iteration is unchanged. As shown in Table~\ref{tab:ssl_4ch}, four-channel pretraining consistently improves both ChannelViT and ChanSFormer over the two-channel setting, and ChanSFormer remains substantially better than ChannelViT in every evaluation setting.}

\begin{table}[!htb]
\CMcolor
\caption{\textbf{Number of channels per view in DINOv2 pretraining} on JUMP-CP Flu., evaluated as in Table~\ref{tab:ssl}. The image batch size is halved for four channels to keep the number of channel views per iteration fixed.}
\label{tab:ssl_4ch}
\centering
\small
\setlength{\tabcolsep}{4pt}
\begin{tabular}{lcccc}
\toprule
 & & CHAMMI & \multicolumn{2}{c}{JUMP-CP} \\
\cmidrule(lr){3-3} \cmidrule(lr){4-5}
Method & \# Channels per view & Avg & Full & Flu. \\
\midrule
\multirow{2}{*}{ChannelViT~\cite{bao2023channel_cvit}} & 2 & 32.88 & 30.72 & 32.97 \\
                                                       & 4 & 34.09 & 31.97 & 35.53 \\
\midrule
\multirow{2}{*}{ChanSFormer} & 2 & 42.86 & 46.13 & 45.15 \\
                             & 4 & \textbf{45.04} & \textbf{48.18} & \textbf{46.49} \\
\bottomrule
\end{tabular}
\end{table}

\FloatBarrier
\section{\texorpdfstring{\CM{Adapting ChannelViT's channel embeddings to unseen channels}}{Adapting ChannelViT's channel embeddings to unseen channels}}
\label{sec:supp:ce_adapt}
\CM{In Table~\ref{tab:ssl}, ChannelViT handles channels that are absent from pretraining by zero-padding their channel embeddings. A stronger baseline is to learn embeddings for the unseen channels on the target dataset. We therefore train the embeddings of the unseen channels together with the linear head and keep all other pretrained weights fixed. This is considerably more expensive than linear probing: computing gradients for the embeddings requires full forward and backward passes through the backbone, whereas linear probing trains a linear head on features extracted once, and several probe configurations can share the same cached features. One adaptation run is about $20\times$ slower than a linear-probe run in our implementation, and each hyperparameter configuration requires a separate run. As shown in Table~\ref{tab:ce_adapt}, adapting the channel embeddings slightly improves ChannelViT over zero-padding, but it remains below ChanSFormer in all seven unseen-channel settings. We omit in-domain settings, where the pretraining and evaluation channels are identical and no embedding needs to be adapted, and CHAMMI as a target, since its official protocol uses feature-space 1-NN classification without trainable downstream parameters.}

\begin{table}[!htb]
\CMcolor
\caption{\textbf{Zero-padded vs.\ adapted channel embeddings for ChannelViT}~\cite{bao2023channel_cvit} on unseen-channel transfer. Models are pretrained with DINOv2 on the \emph{SSL source} and evaluated by linear probe on the \emph{Target}.}
\label{tab:ce_adapt}
\centering
\small
\setlength{\tabcolsep}{4pt}
\begin{tabular}{llccc}
\toprule
SSL source & Target & ChannelViT (zero-pad) & ChannelViT (CE adapt.) & ChanSFormer \\
\midrule
\multirow{4}{*}{CHAMMI}
  & JUMP-CP Full  & 23.94 & 25.54 & \textbf{40.82} \\
  & JUMP-CP Flu.  & 24.71 & 24.99 & \textbf{39.14} \\
  & So2Sat Full   & 66.65 & 66.91 & \textbf{68.08} \\
  & So2Sat Sen-1  & 44.54 & 45.14 & \textbf{46.23} \\
\midrule
\multirow{3}{*}{JUMP-CP Flu.}
  & JUMP-CP Full  & 30.72 & 31.10 & \textbf{46.13} \\
  & So2Sat Full   & 65.00 & 65.19 & \textbf{66.33} \\
  & So2Sat Sen-1  & 44.47 & 44.60 & \textbf{45.34} \\
\bottomrule
\end{tabular}
\end{table}

\FloatBarrier
\section{\texorpdfstring{\CM{Relation to factorized space--time attention}}{Relation to factorized space-time attention}}
\label{sec:supp:factorized}

\CM{Our channel--spatial factorization is related to the space--time factorization of TimeSformer~\cite{bertasius2021space_timesformer}, but serves a different purpose. TimeSformer factorizes attention mainly for efficient video modeling, whereas we use the factorization to remove the dependence on fixed channel embeddings while still allowing information to flow across channels. Moreover, unlike video frames, whose temporal order is meaningful, the order of microscopy channels is arbitrary, so the encoder should be permutation-equivariant along the channel axis and the image-level prediction permutation-invariant. ChanSFormer satisfies both by construction: all operations are shared across channels and no channel ordering is encoded, so channel identity is preserved structurally through the explicit channel axis rather than through learned channel embeddings, and the attention pooling over the channel CLS tokens does not depend on their order. The two designs also differ in how the per-slice representations are aggregated. TimeSformer copies a single CLS token across frames and merges it, which suits highly correlated slices; with the channel CLS tokens replaced by such a shared token, accuracy drops by 1.59\%--3.78\% under the same HCS sampling (Table~\ref{tab:ablation_design}), as averaging the heterogeneous channels of microscopy images is lossy.}

\FloatBarrier
\section{\texorpdfstring{\CM{Learned channel relations}}{Learned channel relations}}
\label{sec:supp:channel_relations}

\begin{figure}[!htb]
\CMcolor
\centering
\includegraphics[width=0.44\linewidth]{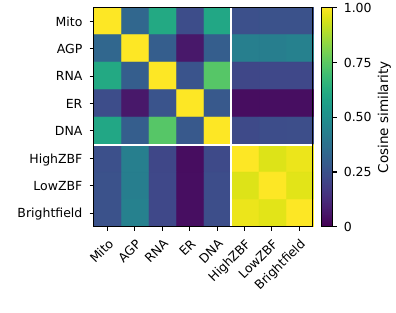}
\hfill
\includegraphics[width=0.44\linewidth]{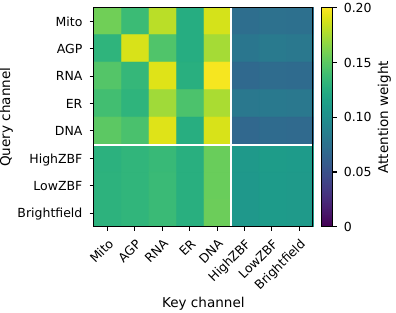}
\caption{\textbf{Learned channel relations on JUMP-CP.} \textbf{Left:} the EMA channel-similarity matrix $S$ used by our sampler. \textbf{Right:} channel-attention weights (row: query, column: key) at patch positions, averaged over all heads, all 12 layers, and the full test set with all 8 channels present; each row sums to 1 (uniform $=0.125$). White lines separate the fluorescence (Mito, AGP, RNA, ER, DNA) and brightfield (HighZBF, LowZBF, Brightfield) channels.}
\label{fig:channel_relations}
\end{figure}

\begin{figure}[!htb]
\CMcolor
\centering
\includegraphics[width=0.44\linewidth]{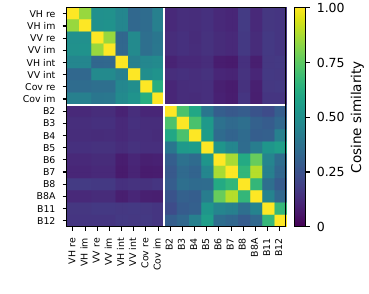}
\hfill
\includegraphics[width=0.44\linewidth]{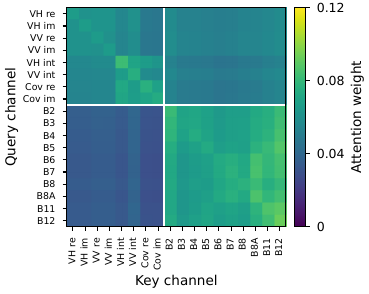}
\caption{\textbf{Learned channel relations on So2Sat.} \textbf{Left:} the EMA channel-similarity matrix $S$ used by our sampler. \textbf{Right:} channel-attention weights (row: query, column: key) at patch positions, averaged over all heads, all 12 layers, and the full test set with all 18 channels present; each row sums to 1. White lines separate the 8 Sentinel-1 channels from the 10 Sentinel-2 bands. Sentinel-1: real (re) and imaginary (im) parts of the unfiltered VH and VV channels, intensity (int) of the refined-Lee-filtered VH and VV channels, and real and imaginary parts of the refined-Lee-filtered covariance-matrix off-diagonal element (Cov).}
\label{fig:channel_relations_so2sat}
\end{figure}

\CM{Figure~\ref{fig:channel_relations} visualizes, on JUMP-CP, the channel-similarity matrix $S$ learned by our sampler (Section~\ref{sec:method:sampling}) and the channel-attention weights of the trained model. $S$ shows a clear two-block structure. The three brightfield channels are nearly interchangeable (pairwise similarity $0.95$--$0.97$), whereas their similarity to the fluorescence channels is substantially lower. The channel attention reflects this structure mainly through the fluorescence queries, which attend mainly to other fluorescence channels and assign each brightfield channel only about $0.07$ of their attention, below the uniform level of $0.125$, whereas the brightfield queries attend nearly uniformly across channels. This is qualitatively consistent with the imaging modalities: fluorescence channels capture complementary stain-specific cellular structures, while the brightfield channels provide a different transmitted-light view. It also agrees with the channel correlation structure reported by ChannelViT~\cite{bao2023channel_cvit}. Although all channel CLS tokens are initialized identically, the learned $S$ is far from uniform, indicating that they do not collapse to the same representation. On So2Sat (Figure~\ref{fig:channel_relations_so2sat}), $S$ likewise shows a two-block structure aligned with the two sensors, Sentinel-1 (SAR) and Sentinel-2 (multispectral), again consistent with ChannelViT.}

%% file: bib.bib
@article{von2021democratising,
  title={Democratising deep learning for microscopy with ZeroCostDL4Mic},
  author={Von Chamier, Lucas and Laine, Romain F and Jukkala, Johanna and Spahn, Christoph and Krentzel, Daniel and Nehme, Elias and Lerche, Martina and Hern{\'a}ndez-P{\'e}rez, Sara and Mattila, Pieta K and Karinou, Eleni and others},
  journal={Nature communications},
  volume={12},
  number={1},
  pages={2276},
  year={2021},
  publisher={Nature Publishing Group UK London}
}

@article{vincent2022phenotypic,
  title={Phenotypic drug discovery: recent successes, lessons learned and new directions},
  author={Vincent, Fabien and Nueda, Arsenio and Lee, Jonathan and Schenone, Monica and Prunotto, Marco and Mercola, Mark},
  journal={Nature Reviews Drug Discovery},
  volume={21},
  number={12},
  pages={899--914},
  year={2022},
  publisher={Nature Publishing Group UK London}
}

@article{boutros2015microscopy,
  title={Microscopy-based high-content screening},
  author={Boutros, Michael and Heigwer, Florian and Laufer, Christina},
  journal={Cell},
  volume={163},
  number={6},
  pages={1314--1325},
  year={2015},
  publisher={Elsevier}
}

@article{bray2016cell,
  title={Cell Painting, a high-content image-based assay for morphological profiling using multiplexed fluorescent dyes},
  author={Bray, Mark-Anthony and Singh, Shantanu and Han, Han and Davis, Chadwick T and Borgeson, Blake and Hartland, Cathy and Kost-Alimova, Maria and Gustafsdottir, Sigrun M and Gibson, Christopher C and Carpenter, Anne E},
  journal={Nature protocols},
  volume={11},
  number={9},
  pages={1757--1774},
  year={2016},
  publisher={Nature Publishing Group UK London}
}

@article{chen2023chammi_depthvit,
  title={CHAMMI: A benchmark for channel-adaptive models in microscopy imaging},
  author={Chen, Zitong Sam and Pham, Chau and Wang, Siqi and Doron, Michael and Moshkov, Nikita and Plummer, Bryan and Caicedo, Juan C},
  journal={Advances in Neural Information Processing Systems},
  volume={36},
  pages={19700--19713},
  year={2023}
}

@inproceedings{kraus2024masked_camae,
  title={Masked autoencoders for microscopy are scalable learners of cellular biology},
  author={Kraus, Oren and Kenyon-Dean, Kian and Saberian, Saber and Fallah, Maryam and McLean, Peter and Leung, Jess and Sharma, Vasudev and Khan, Ayla and Balakrishnan, Jia and Celik, Safiye and others},
  booktitle={Proceedings of the IEEE/CVF Conference on Computer Vision and Pattern Recognition},
  pages={11757--11768},
  year={2024}
}

@article{bao2023channel_cvit,
  title={Channel vision transformers: an image is worth 1 x 16 x 16 words},
  author={Bao, Yujia and Sivanandan, Srinivasan and Karaletsos, Theofanis},
  journal={arXiv preprint arXiv:2309.16108},
  year={2023}
}

@inproceedings{bourriez2024chada_chadavit,
  title={Chada-vit: Channel adaptive attention for joint representation learning of heterogeneous microscopy images},
  author={Bourriez, Nicolas and Bendidi, Ihab and Cohen, Ethan and Watkinson, Gabriel and Sanchez, Maxime and Bollot, Guillaume and Genovesio, Auguste},
  booktitle={Proceedings of the IEEE/CVF Conference on Computer Vision and Pattern Recognition},
  pages={11556--11565},
  year={2024}
}

@article{pham2024enhancing_dichavit,
  title={Enhancing feature diversity boosts channel-adaptive vision transformers},
  author={Pham, Chau and Plummer, Bryan},
  journal={Advances in Neural Information Processing Systems},
  volume={37},
  pages={89782--89805},
  year={2024}
}

@inproceedings{bertasius2021space_timesformer,
  title={Is space-time attention all you need for video understanding?},
  author={Bertasius, Gedas and Wang, Heng and Torresani, Lorenzo},
  booktitle={Proceedings of the 38th International Conference on Machine Learning (ICML)},
  series={Proceedings of Machine Learning Research},
  volume={139},
  pages={813--824},
  year={2021}
}

@article{dosovitskiy2020image_vit,
  title={An image is worth 16x16 words: Transformers for image recognition at scale},
  author={Dosovitskiy, Alexey and Beyer, Lucas and Kolesnikov, Alexander and Weissenborn, Dirk and Zhai, Xiaohua and Unterthiner, Thomas and Dehghani, Mostafa and Minderer, Matthias and Heigold, Georg and Gelly, Sylvain and others},
  journal={arXiv preprint arXiv:2010.11929},
  year={2020}
}

@article{chandrasekaran2024three_jumpcp,
  title={Three million images and morphological profiles of cells treated with matched chemical and genetic perturbations},
  author={Chandrasekaran, Srinivas Niranj and Cimini, Beth A and Goodale, Amy and Miller, Lisa and Kost-Alimova, Maria and Jamali, Nasim and Doench, John G and Fritchman, Briana and Skepner, Adam and Melanson, Michelle and others},
  journal={Nature Methods},
  volume={21},
  number={6},
  pages={1114--1121},
  year={2024},
  publisher={Nature Publishing Group US New York}
}

@article{bao2023contextual_jumpcpdata,
  title={Contextual vision transformers for robust representation learning},
  author={Bao, Yujia and Karaletsos, Theofanis},
  journal={arXiv preprint arXiv:2305.19402},
  year={2023}
}

@inproceedings{he2022masked,
    title={Masked autoencoders are scalable vision learners},
    author={He, Kaiming and Chen, Xinlei and Xie, Saining and Li, Yanghao and Doll{\'a}r, Piotr and Girshick, Ross},
    booktitle={Proceedings of the IEEE/CVF conference on computer vision and pattern recognition},
    pages={16000--16009},
    year={2022}
}

@article{zhou2021ibot,
  title={ibot: Image bert pre-training with online tokenizer},
  author={Zhou, Jinghao and Wei, Chen and Wang, Huiyu and Shen, Wei and Xie, Cihang and Yuille, Alan and Kong, Tao},
  journal={arXiv preprint arXiv:2111.07832},
  year={2021}
}

@inproceedings{caron2021emerging,
  title={Emerging Properties in Self-Supervised Vision Transformers},
  author={Caron, Mathilde and Touvron, Hugo and Misra, Ishan and J\'egou, Herv\'e  and Mairal, Julien and Bojanowski, Piotr and Joulin, Armand},
  booktitle={Proceedings of the International Conference on Computer Vision (ICCV)},
  year={2021}
}

@article{moutakanni2025cell,
  title={Cell-DINO: Self-supervised image-based embeddings for cell fluorescent microscopy},
  author={Moutakanni, Th{\'e}o and Couprie, Camille and Yi, Seungeun and Doron, Michael and Chen, Zitong S and Moshkov, Nikita and Gardes, Elouan and Caron, Mathilde and Touvron, Hugo and Joulin, Armand and others},
  journal={PLOS Computational Biology},
  volume={21},
  number={12},
  pages={e1013828},
  year={2025},
  publisher={Public Library of Science San Francisco, CA USA}
}

@article{de2025scaling,
  title={Scaling channel-adaptive self-supervised learning},
  author={De Lorenci, Alice V and Yi, Seung Eun and Moutakanni, Th{\'e}o and Bojanowski, Piotr and Couprie, Camille and Caicedo, Juan C and Pernice, Wolfgang Maximilian Anton},
  journal={Transactions on Machine Learning Research},
  year={2025}
}

@article{lian2025isolated,
  title={Isolated channel vision transformers: from single-channel pretraining to multi-channel finetuning},
  author={Lian, Wenyi and Micke, Patrick and Lindblad, Joakim and Sladoje, Nata{\v{s}}a},
  journal={arXiv preprint arXiv:2503.09826},
  year={2025}
}

@article{oquab2023dinov2,
  title={Dinov2: Learning robust visual features without supervision},
  author={Oquab, Maxime and Darcet, Timoth{\'e}e and Moutakanni, Th{\'e}o and Vo, Huy and Szafraniec, Marc and Khalidov, Vasil and Fernandez, Pierre and Haziza, Daniel and Massa, Francisco and El-Nouby, Alaaeldin and others},
  journal={arXiv preprint arXiv:2304.07193},
  year={2023}
}

@article{dao2022flashattention,
  title={Flashattention: Fast and memory-efficient exact attention with io-awareness},
  author={Dao, Tri and Fu, Dan and Ermon, Stefano and Rudra, Atri and R{\'e}, Christopher},
  journal={Advances in neural information processing systems},
  volume={35},
  pages={16344--16359},
  year={2022}
}

@article{dao2023flashattention,
  title={Flashattention-2: Faster attention with better parallelism and work partitioning},
  author={Dao, Tri},
  journal={arXiv preprint arXiv:2307.08691},
  year={2023}
}

@Misc{xFormers2022,
  author =       {Benjamin Lefaudeux and Francisco Massa and Diana Liskovich and Wenhan Xiong and Vittorio Caggiano and Sean Naren and Min Xu and Jieru Hu and Marta Tintore and Susan Zhang and Patrick Labatut and Daniel Haziza and Luca Wehrstedt and Jeremy Reizenstein and Grigory Sizov},
  title =        {xFormers: A modular and hackable Transformer modelling library},
  howpublished = {\url{https://github.com/facebookresearch/xformers}},
  year =         {2022}
}

@InProceedings{plummerNPAS2022,
  author={Bryan A. Plummer and Nikoli Dryden and Julius Frost and Torsten Hoefler and Kate Saenko},
  title={Neural Parameter Allocation Search},
  booktitle={International Conference on Learning Representations (ICLR)},
  year={2022}
}

@inproceedings{savarese2018learning,
  title={Learning Implicitly Recurrent {CNN}s Through Parameter Sharing},
  author={Pedro Savarese and Michael Maire},
  booktitle={International Conference on Learning Representations (ICLR)},
  year={2019}
}

@inproceedings{haHypernetworks2016,
  title={HyperNetworks},
  author={David Ha and Andrew Dai and Quoc Le},
  booktitle={International Conference on Learning Representations (ICLR)},
  year={2016}
}

@inproceedings{phamcha,
  title={ChA-MAEViT: Unifying Channel-Aware Masked Autoencoders and Multi-Channel Vision Transformers for Improved Cross-Channel Learning},
  author={Pham, Chau and Caicedo, Juan C and Plummer, Bryan A},
  booktitle={The Thirty-ninth Annual Conference on Neural Information Processing Systems},
  year={2025}
}

@article{zhu2020so2sat,
  title={So2Sat LCZ42: A benchmark data set for the classification of global local climate zones [software and data sets]},
  author={Zhu, Xiao Xiang and Hu, Jingliang and Qiu, Chunping and Shi, Yilei and Kang, Jian and Mou, Lichao and Bagheri, Hossein and Haberle, Matthias and Hua, Yuansheng and Huang, Rong and others},
  journal={IEEE Geoscience and Remote Sensing Magazine},
  volume={8},
  number={3},
  pages={76--89},
  year={2020},
  publisher={IEEE}
}

@inproceedings{miao2025multiview,
  title={Multi-view Gaze Target Estimation},
  author={Miao, Qiaomu and Golani, Vivek Raju and Xu, Jingyi and Dutta, Progga Paromita and Hoai, Minh and Samaras, Dimitris},
  booktitle={Proceedings of the IEEE/CVF International Conference on Computer Vision (ICCV)},
  pages={5371--5381},
  year={2025}
}
